\documentclass[conference]{IEEEtran}
\usepackage{cite}
\usepackage{amsmath,amssymb,amsfonts}
\usepackage{algorithmic}
\usepackage{graphicx}
\usepackage{textcomp}
\usepackage{xcolor}
\usepackage{soul}
\usepackage{extdash}
\def\BibTeX{{\rm B\kern-.05em{\sc i\kern-.025em b}\kern-.08em
    T\kern-.1667em\lower.7ex\hbox{E}\kern-.125emX}}

\usepackage{tikz}
\usetikzlibrary{positioning}
\usetikzlibrary{calc}
\usetikzlibrary{shapes.misc}

\usepackage{ulem}
\usepackage{comment}
\usepackage{subfigure}

\begin{document}

\title{Active Gaze Control for Foveal Scene Exploration}

\author{\IEEEauthorblockN{Alexandre M. F. Dias, Luís Simões, Plinio Moreno, Alexandre Bernardino}
\IEEEauthorblockA{\textit{Instituto Superior Técnico}\\
Lisbon, Portugal \\
\{alexandre.f.dias, luis.d.simoes, alexandre.bernardino\}@tecnico.ulisboa.pt, plinio@isr.tecnico.ulisboa.pt}
}

\maketitle

\begin{abstract}
Active perception and foveal vision are the foundations of the human visual system. While foveal vision reduces the amount of information to process during a gaze fixation, active perception will change the gaze direction to the most promising parts of the visual field. We propose a methodology to emulate how humans and robots with foveal cameras would explore a scene, identifying the objects present in their surroundings with in least number of gaze shifts. Our approach is based on three key methods.
First, we take an off-the-shelf deep object detector, pre-trained on a large dataset of regular images, and calibrate the classification outputs to the case of foveated images.
Second, a body-centered semantic map, encoding the objects classifications and corresponding uncertainties, is sequentially updated with the calibrated detections, considering several data fusion techniques.
Third, the next best gaze fixation point is determined based on information-theoretic metrics that aim at minimizing the overall expected uncertainty of the semantic map. When compared to the random selection of next gaze shifts, the proposed method achieves an increase in detection F1-score of 2-3 percentage points for the same number of gaze shifts and reduces to one third the number of required gaze shifts to attain similar performance.

\end{abstract}


\section{Introduction}

\begin{figure}[htbp!]
\centering
\includegraphics[width=0.49\linewidth]{images/283322282_c7c58460cc_z.jpg}
\includegraphics[width=0.49\linewidth]{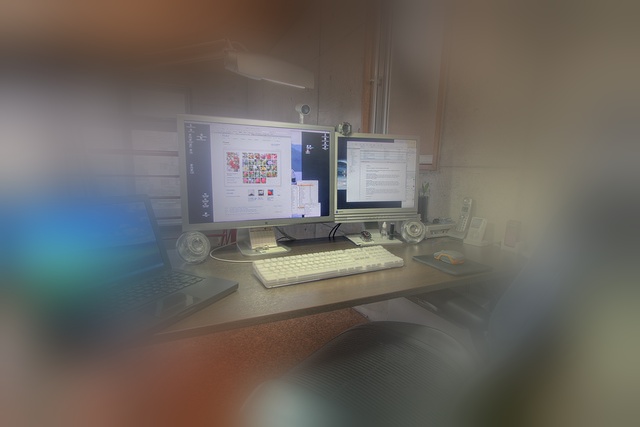}
\caption{Regular image (left) and corresponding foveal representation (right) with the fovea in the center of the image.}
\label{fig:cart_fov}
\end{figure}

The fovea is the area of the visual retina with the highest concentration of photoreceptors. In humans, the  fovea comprises less than 1\% of retinal size, but takes up over 50\% of the visual cortex \cite{Krantz2012}. In artificial systems, a foveated image is an image having an area of high resolution (typically the center), where fine details of objects can be observed, while the remaining of the image (periphery) has low resolution and objects look blurry. An illustration of an artificial foveal image is shown in Fig. \ref{fig:cart_fov}.

Central or foveal vision is an indispensable feature of the human eye, allowing activities that require high visual acuity such as reading, sports and driving to be carried out. It has been implemented with success in robots to perform visual tasks such as object tracking and depth perception with low computational resources \cite{JAVIERTRAVER2010378}. Visual search tasks require nonetheless the involvement of peripheral vision, from which it has proved challenging for image processing and pattern recognition methods to extract relevant information.

Although there has been a large amount of research and developments on visual attention and visual search models \cite{Aydemir2011SearchIT,Aydemir2013,Almeida2018,Melicio}, there is still a long way to go, specially regarding the modeling of the mechanisms that help the decision of where to shift the gaze to. Some approaches consider interesting points to look at as points that are salient in some image dimension (e.g. color) or have high uncertainty in other dimensions (e.g. depth) \cite{Grotz,Figueiredo}, but lack on the semantic representation of the observed objects. Other approaches make use of recent deep object detectors, that carry semantic information about the classes of objects, to detect objects of interest in the periphery of the visual field and perform visual search tasks \cite{Almeida2018,Melicio}. However, a major challenge in visual search with foveal images is the fact that object detectors and saliency models have not been designed to work on low resolution images, thus lagging behind in performance in the detection of objects in the peripheral regions of a foveal image.

In this paper, we propose a methodology that allows exploring a foveal scene while minimizing the required number of gaze fixations in order to identify the objects present in the scene. Scene exploration is achieved by using a readily available deep object detector and by implementing three key elements: a Foveal Observation Model, which learns the response of the Object Detector to foveated images; Data Fusion techniques, which combine the information of subsequent detections in order to update the knowledge about the scene; Active Perception techniques, which determine the best location in the scene to where the gaze shall be shifted. To the extent of our knowledge, this is the first attempt at combining Object Detection and Active Perception to perform scene exploration with foveal vision.

The paper is structured as follows: in Section \ref{sec:proposed_methodology}, the proposed methodology and its key elements are justified and presented in detail; in Section \ref{sec:results_discussion}, the experimental setup is described and the results of the experiments are presented and discussed; in Section \ref{sec:concl}, we conclude regarding the validity of the proposed methods.

\vfill

\section{Proposed Methodology}
\label{sec:proposed_methodology}

\subsection{Overview}
\label{subsec:prob_def}
The main goal of this work is to implement a methodology optimizing the exploration of a scene using foveal vision, to gather as much information as possible about the objects present in the scene in the least number of gaze shifts.

\begin{figure}[!bp]
\centering
\includegraphics[width=\linewidth]{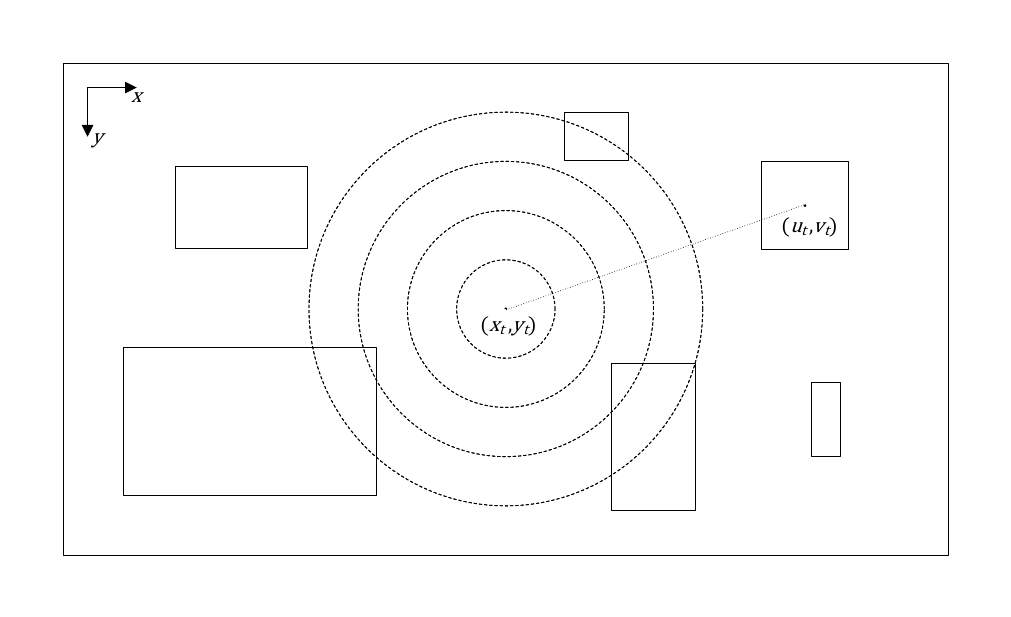}
\caption{Scene representation: world coordinates $x,y$ and local image coordinates $u,v$; the rectangles correspond to the objects present and the center of the dashed circles is the focal point $(x_t,y_t)$.}
\label{fig:prob_scheme}
\end{figure}

We assume the observing agent to be stationary, analogous to a human not moving the head nor the body, yet to be able to, without any uncertainty, i) shift its gaze to certain parts of the visual field and ii) sense the current gaze direction. The transformation from world to retinal coordinates is defined as a simple translation. The agent's gaze being fixed at world coordinates $(x_t, y_t)$, an observation at time step $t$ in world coordinates $(x, y)$ corresponds to an observation in local image coordinates $(u_t, v_t) = (x-x_t, y-y_t)$, and the distance $d_t$ to the focal point is given by $d_t = \left\|\left(u_t, v_t\right)\right\|$ --- the subscript of $d_t$ will be omitted for simplicity, unless required to avoid ambiguity. In Fig.~\ref{fig:prob_scheme}, a schematic illustration of the spatial scene representation considered is shown.

After the fixation of the gaze at time step $t$, the image is foveated. The foveation method takes into account the current focal point (center of the fovea) and foveates the full image such that the foveal region keeps the spectral content of the original image and the regions in the periphery are low-pass filters with decreasing bandwidth as the distance from the fovea increases. The used foveation method is described in \cite{Almeida2018,Melicio} and code is available online\footnote{\scriptsize \texttt{github.com/vislab-tecnico-lisboa/laplacian-foveation}}.

\begin{figure}[!bp]
\centering
\includegraphics[width=0.75\linewidth]{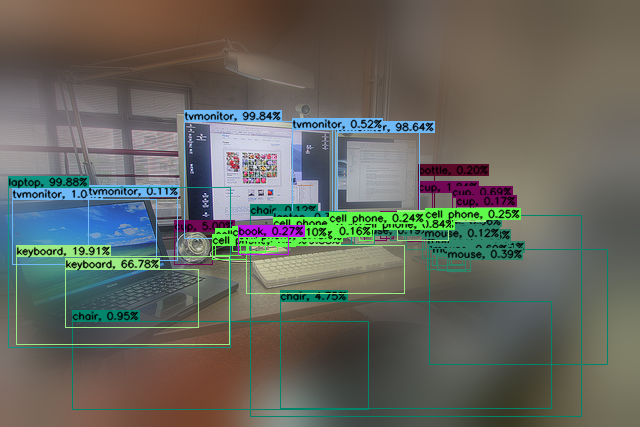}
\caption{Object detections in a foveated image. The names and percentages indicate the class with the highest confidence score of the respective detection (bounding box).}
\label{fig:foveated1}
\end{figure}

The foveated image then serves as input to an objector detector,  which computes a set of $L_t$ detections, denoted as ${\cal I}_t = \{I_{t}^{l_t}\}, l_t = 1,\ldots,L_t$, where $I_{t}^{l_t} = ({B}_{t}^{l_t}, {S}_{t}^{l_t})$ is a 2-tuple comprised of the location and size (bounding box) of the $l_{t}$-th detection, ${B}_{t}^{l_t}$, and the respective class confidence scores, ${S}_{t}^{l_t}~=~\left(s_{t,0}^{l_t}, s_{t,1}^{l_t}, \ldots, s_{t,K}^{l_t}\right) \in~\Delta^{K}$. The set of possible object classes is ${\cal C} = \{0, 1,\ldots, K\}$,
where $K$ is the number of object classes and $0$ denotes the background class, i.e., a class representing the nonexistence of an object. In Fig. \ref{fig:foveated1}, object detections in a foveated image are depicted: the names and percentages correspond to the class of each detection having the highest confidence score.

Nevertheless, off-the-shelf object detectors do not provide classification scores appropriate for foveated images. As they are trained on regular images, for objects undistorted by foveal vision (objects in the foveal region) the detector output is expected to be similar to that of a regular image; however, for objects in the periphery of the fovea, the detector is expected to output classification scores with increased uncertainty (increased entropy) due to the additional class ambiguity in the blurred part of the visual field. Thus, a Foveal Observation Model learns how the class confidence scores of the detections vary with the position relative to the focal point, allowing to calibrate such scores and providing uncertainty measures of the detections. This approach sidesteps the retraining of the object detector's neural network, which can be excessively time consuming.

On the other hand, a coarse detection in a foveated image corresponding to the peripheral view of an object will inform the agent of the potential semantic classes that may be in that location and should prompt the gaze to be shifted to that location if the information about the object is still scant, i.e., so far there has been no gaze fixation close to the object. A semantic map would then be improved by a gaze fixation closer to such spacial location. Then, to perform scene exploration efficiently, the detector output, after calibration by the observation model, is merged with previous detection information to update a body-centered semantic map with the knowledge about the scene.

Active Perception can then exploit this updated knowledge to determine the next best viewpoint, i.e., the focal point producing new observations expected to decrease the overall uncertainty about the scene. The gaze is then shifted towards this point, and the image is again foveated to reflect this new choice of focal point. This process of image foveation, object detection, map information update and next best gaze fixation determination is then repeated until a suitable stopping criterion is met.

\tikzset{
    -|/.style={to path={-| (\tikztotarget)}},
    |-/.style={to path={|- (\tikztotarget)}},
}

\begin{figure*}[!htbp]
\centering
\begin{tikzpicture} [
   node distance = 14mm, 
   font=\scriptsize, 
   functionbox/.style={rectangle, thick, draw=black, minimum height = 7mm, minimum width = 7mm, text centered}, 
]
\node (input) [functionbox, text width=10mm]{Image dataset};
\node (fovea) [functionbox, right=of input, text width=12mm]{Foveal Conversion};
\node (yolo)   [functionbox, right=of fovea, text width=12mm]{Object Detector};
\node (obs)    [functionbox, right=of yolo, text width=12mm]{Foveal Observation Model};
\node (fusion) [functionbox, right=of obs, text width=10mm]{Data Fusion};
\node (active) [functionbox, right=of fusion, text width=12mm]{Active Perception};
\node (gaze)   [functionbox, right=of active, text width=8mm]{Gaze Shift};
\path (input) edge[->] node [above, font=\scriptsize, text width=10mm] {selected image} (fovea);
\path (fovea) edge[->] node [above, font=\scriptsize, text width=10mm] {foveated image} (yolo);
\path (yolo)  edge[->] node [above, font=\scriptsize, text width=10mm] {detected objects} (obs);
\path (obs)   edge[->] node [above, font=\scriptsize, text width=10mm] {calibrated scores} (fusion);
\path (fusion) edge[->] node [above, font=\scriptsize, text width=10mm] {updated map} (active);
\path (active) edge[->] node [above, font=\scriptsize, text width=12mm] {next best gaze point} (gaze);
\draw [->] (gaze.north) -- ++ (0.0,0.6) node [above, font=\scriptsize, text width=10mm] {current viewpoint} -| (obs.north);
\draw [->] (gaze.north) -- ++ (0.0,0.6) -| (fovea.north);
\end{tikzpicture}
\caption[Components of the proposed methodology.]{Components of the proposed methodology.}
\label{fig:methodology_components}
\end{figure*}
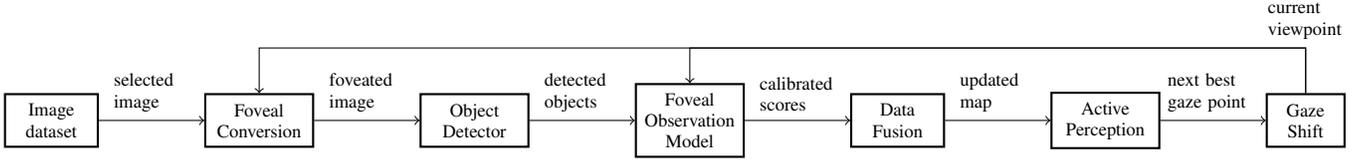

\subsection{Foveal Observation Model}
\label{subsec:foveal_observation_model}

The Foveal Observation Model should account for the increased uncertainty of each detection due to the foveation process. The detector output scores, $S_{t}^{l_t}$, when normalized to unit sum, can be seen as a realization of a Dirichlet distribution. Thus, for a given labeled dataset, the parameters of the Dirichlet distributions that characterize the objector detector response to each object class $k\in{\cal C}$ at each distance $d$ to the fovea, ${\boldsymbol{\alpha}_{k,d} =  \left( {\alpha_{k,d,1}, \alpha_{k,d,2}, \ldots, \alpha_{k,d,K}} \right)}$, can be estimated. The Foveal Observation Model thus consists of sets of Dirichlet distribution parameters --- one set of parameter values per object class $k$ and per distance level $d$ (7 different distance levels were considered), and it is trained to learn the probability distribution $p( S | C=k, d )$.

The normalized scores vector $S_{t}^{l_t}$ corresponding to each new detection $I_{t}^{l_t}$ being modeled as realizations of the previously trained Dirichlet distributions: $S \sim \textrm{Dir}(\boldsymbol{\alpha}_{k,d}$), for each class $k=1,...,K$, a new set of calibrated classification scores $S'$ can be computed that more accurately reflect the probabilities of detections in the foveal images:
\begin{equation}
\begin{aligned}
    {S}' &= (s'_{1}, \ldots, s'_{K})
    \\
    &= \frac{1}{D}\left[p({S_{t}^{l_t}}|C=1,d), ..., p({S_{t}^{l_t}}|C=K,d)\right],
    \label{eq:obsModel_scores}
\end{aligned}
\end{equation}
where $D=\sum_{k=1}^K p({S_{t}^{l_t}}|C=k,d)$ is a normalization factor such that $\sum_{k=1}^K s'_{k}=1$. The calibrated classification scores ${S}'$ are expected to have less entropy than the original $S$.

\subsection{Data Fusion}
\label{sec:fusion_model}

The fusion problem consists in, given a set of classification scores for a single pattern, calculating an overall classification score $\mathbf{p}$ or estimating a distribution of classification scores. As the agent explores the scene by shifting the gaze to different parts of the visual field, the information of subsequent detections is collected. This information is used to update a map $\mathbf{M}$. The $\left(K+1\right)$-channel map $\mathbf{M}_t(x,y)$ is defined as:
\begin{equation*}
    \mathbf{M}_{t}(x,y) = \mathbf{p}_{t}^{x,y} = (p_{t,0}^{x,y},p_{t,1}^{x,y},\ldots,p_{t,K}^{x,y}) \in \Delta^{K},
    \label{eq:fusion_model}
\end{equation*}
where $p_{t,k}^{x,y} = P(C^{x,y} = k | {\cal I}_{1:t}, x_{1:t}, y_{1:t})$ specifies the class posterior probability of class $k \in {\cal C}$. Thus, $\mathbf{p}_{t}^{x,y}$ constitutes the estimate, at time step $t$, of the class posterior probabilities of an object located at coordinates $(x,y)$.

Given a sequence of classifier scores ${S}_{1:t}$ for time steps 1 up to $t$, world coordinates $(x,y)$ and distances $d_{1:t}$ to the fovea, a simple product rule (Na\"{i}ve Bayes) can be used to update the map information, according to 
\begin{equation}
\begin{aligned}
    {M}_{t,k}({x,y}) &= P(C^{x,y}=k|{S}_{1:t},d_{1:t})
    \\&= \frac{1}{Z}\prod_{\tau=1}^t p({S}_{\tau}|C^{x,y}=k) \\&=\frac{1}{Z_{t}}p({S}_t|C^{x,y}=k) M_{t-1,k}({x,y})
\end{aligned}
\label{eq:prod_rule}
\end{equation}
where $Z$ and $Z_{t}$ are unit-sum normalizing constants.

A categorical distribution is unable to deal with the uncertainty in the estimated parameters $\mathbf{p}$ (variance), only with the first moment (expected value). Such uncertainty can, however, be captured by modeling the parameters $\mathbf{p}$ as realizations of a Dirichlet distribution, $\mathbf{p}~\sim~\textrm{Dir}\left(\boldsymbol{\beta}\right)$, $\boldsymbol{\beta}=\left(\beta_0,\ldots,\beta_K\right)$.
This leads to the definition of a map $\mathbf{B}_t(x,y)$ containing the estimates at time step $t$ and for world coordinates $\left(x,y\right)$ of the parameters:
\begin{equation}
\mathbf{B}_{t}({x,y}) = \boldsymbol{\beta}_{t}^{x,y} = \left(\beta_0^{x,y}~,\beta_1^{x,y},\ldots,\beta_K^{x,y}\right)
\label{eq:dir_map}
\end{equation}

These parameters are able to encode the uncertainty --- through the second moment or variance --- of the class posterior scores $p^{x,y}_{t,k}$ and allow for practical sequential update rules. The class posterior probabilities of $\mathbf{M}_t(x,y)$ can be computed from the parameters of $\mathbf{B}_t(x,y)$ through 
\begin{equation}
    {M}_{t,k}({x,y}) = p^{x,y}_{t,k} = \mathbb{E}_{p_{t,k}^{x,y}}\left[\textrm{Dir}\left(\boldsymbol{\beta}_{t}^{x,y}\right) \right] = \frac{\beta^{x,y}_{t,k}}{\sum_{j=1}^K\beta^{x,y}_{t,j}}.
    \label{eq:expected_value}
\end{equation}

Several strategies for the sequential update of the map $\mathbf{M}_t$ exist. In \cite{6187195}, Kaplan \textit{et al.} proposes several update rules for the fusion of classifier scores: the product rule (equivalent to  Na\"{i}ve Bayes presented), the sum rule, and a novel approach that we denote by Kaplan's rule. 

The application of the sum rule consists in simply adding each of the classifier scores to the parameter of the corresponding class $k$:
\begin{equation}
    \beta_{t+1,k} = \beta_{t,k} + s_{t,k}.
\label{eq:naive_update}
\end{equation}
Differently from the product rule, in the sum rule the observations are summed instead of multiplied, providing robustness to outliers. On the other hand, the sum rule does not yield a posterior Dirichlet distribution.

In Kaplan's rule \cite{6187195}, a Dirichlet distribution is fitted to the actual posterior through a moment matching approach. Nevertheless, using this rule with the raw output of the detector would ignore the knowledge about the influenced of the objects location relative to the center of the fovea (Foveal Observation Model). Thus, in order to assess a potential improvement in exploring the scene, a modified version of Kaplan's rule can consider the calibrated scores ${S}'_{t}$ of (\ref{eq:obsModel_scores}) instead of the original detector scores ${S}_{t}$:

\begin{equation}
    \beta_{t+1,k} = \frac{\beta_{t,k}\left(1 + \frac{s'_{t,k}}{\sum^K_{j=1}\beta_{t,j}s'_{t,j}}\right)}{1+\frac{\min_js'_{t,j}}{\sum^K_{j=1}\beta_{t,j}s'_{t,j}}}
    \label{eq:mod_kaplan}
\end{equation}

\subsection{Active Perception}
\label{subsec:active_perception}

In order to perform scene exploration in the least number of gazes, and based on the updated knowledge of the map, an Active Perception method chooses the point $(x_{t+1},y_{t+1})$ to which to shift the gaze in the next time step $t+1$, minimizing some measure of uncertainty $F$:
\begin{equation}
    \left( x^*_{t+1},y^*_{t+1} \right) = \underset{x_{t+1},y_{t+1}}{\operatorname{arg min}} \; F(x_{t+1},y_{t+1},{\cal I}_{1:t}, x_{1:t}, y_{1:t}),
    \label{eq:active_perception}
\end{equation}

This requires predicting the information accumulated up to time step $t+1$, ${\cal I}_{1:t+1}$, that would be obtained if the gaze direction were changed to coordinates $(x',y') = (x^*_{t+1},y^*_{t+1})$. As the knowledge about the scene at time $t$ is represented by the semantic map $\mathbf{B}_t(x,y)$, we need to compute:
\begin{equation}
\bar{\mathbf{B}}_{t+1}^{x',y'}(x,y) = \mathbb{E}[\mathbf{B}_{t+1}(x,y) | \mathbf{B}_{t}(x,y), x', y' ]
\label{eq:expected_map}
\end{equation}

To perform the computation of (\ref{eq:expected_map}) we need to predict the detector classification scores for each world coordinate $(x,y)$, given that the gaze is shifted to $(x',y')$. Let ${d'=||(x-x',y-y')||}$. Then, $p(S^{x,y}|x',y',B_{t,k}^{x,y})$ can be written as:
\begin{equation}
\begin{aligned}
& p(S^{x,y}|x',y',B_{t,k}^{x,y}) = \\
& = \sum_{k=0}^K p(S^{x,y},k|x',y',B_{t,k}^{x,y}) \\
& = \sum_{k=0}^K p(S^{x,y}|k,x',y',B_{t,k}^{x,y}) P(k | x',y',B_{t,k}^{x,y}) \\
&  = \sum_{k=0}^K p(S^{x,y}|k,x',y') P(k | B_{t,k}^{x,y})
\end{aligned}
\end{equation}
where the last step used the fact that $S^{x,y}$ is conditionally independent of $B_{t,k}^{x,y}$ given the object class $k$, and that the object class $k$ at coordinates $(x,y)$ does not depend on the placement of the fovea at time $t+1$.
Finally, since $P(k | B_{t,k}^{x,y}) = \beta_{t,k}^{x,y}/\sum_{i=0}^K \beta_{t,i}^{x,y}$:
\begin{equation*}
p(S^{x,y}|x',y',B_{t,k}^{x,y}) 
\propto \sum_{k=0}^K p(S^{x,y}|k,x',y') \frac{\beta_{t,k}^{x,y}}{\sum_{i=0}^K \beta_{t,i}^{x,y}}.
\end{equation*}
Given the linearity of the expectation operator:
\begin{equation}
\bar{S}^{x,y}=\mathbb{E}(S^{x,y}|x',y')= \sum_{k=0}^K \frac{\alpha_{k,d}}{\sum_{i=0}^K \alpha_{i,d}} \frac{\beta_{t,k}^{x,y}}{\sum_{i=0}^K \beta_{t,i}^{x,y}}.
\end{equation}
The predicted classifier scores can then be used to run the data fusion processes (\ref{eq:prod_rule},\ref{eq:naive_update},\ref{eq:mod_kaplan}) and compute (\ref{eq:expected_map}).

A measure of uncertainty is associated to the class posterior distribution at each map coordinate, represented as $U(x,y)$. Several measures of uncertainty are considered: the Kullback-Leibler (KL) divergence or relative entropy, the entropy, and the difference between the scores of the two classes with highest confidence.

The KL divergence is used to measure the information gain between the current Dirichlet distribution of parameters ${\boldsymbol{\beta}}$ and an initial Dirichlet uniform distribution ($\beta_{0,k}=1, k=0,\ldots,K$):
\begin{equation}
    U_t(x,y) = -D_{KL}(\textrm{Dir} \left(\boldsymbol{\beta}^{x,y}_t \right) || \textrm{Dir}(\boldsymbol{\beta}_{0})).
\label{eq:kl}
\end{equation}
The entropy of a probability distribution measures the degree of randomness in the random variable. The entropy is used as a measure of uncertainty of the Dirichlet distributions of the map:
\begin{equation}
    U_t(x,y) = H(\textrm{Dir}(\boldsymbol{\beta}^{x,y}_t)),\; \forall x,y
\label{eq:entropy}
\end{equation}
The last uncertainty measure is defined as the difference between the two classes of largest expected value. Let $p_{t,i}^{x,y}$ and $p_{t,j}^{x,y}$ be, respectively, the largest and second largest expected class posterior probabilities at map coordinates $(x,y)$. Then, we minimize
\begin{equation}
    U_t(x,y) = -|p_{t,i}^{x,y} - p_{t,j}^{x,y}|,
\label{eq:two-peaks}
\end{equation}

The predictions for the map at time $t+1$ being given by (\ref{eq:expected_map}), the expected values of the uncertainty for all possible next gaze locations $(x',y')$ are computed through: 
\begin{equation}
    U_{t+1}^{x',y'}(x,y) = \mathbb{E}[U_t(x,y) | \bar{B}_{t+1}^{x',y'}(x,y)]
\end{equation}
Finally, we minimize over all possible next gaze directions $(x',y')$ the expected uncertainty accumulated over all map coordinates $(x,y)$:
\begin{equation}
    (x^{*},y^{*}) =  \underset{x',y'}{\operatorname{argmin}} \sum_{x}\sum_{y} U_{t+1}^{x',y'}(x,y) 
\end{equation}

\section{Results and Discussion}
\label{sec:results_discussion}

\subsection{Experimental Setup}

A set of 50 images was randomly chosen from the COCO 2017 validation dataset (80 object classes) as scenes in which to perform visual exploration. The employed Deep Object Detector \cite{redmon2018yolov3} is pre-trained on the COCO 2017 training dataset. To validate our approach, we have incrementally implemented the pipeline illustrated in Fig. \ref{fig:methodology_components}. The incremental implementation corresponds to the successive evaluation of each of the three key methods of the approach:

\begin{enumerate}
    \item Foveal Observation Model: the parameters of each Dirichlet distribution are learned by employing an iterative method \cite{minka}; to analyse the effect of class classification scores calibration by the Foveal Observation Model in a simple classification task, the 50 images were foveated at randomly chosen focal points; the match between the detected bounding boxes and the ground-truth bounding boxes was considered valid for an Intersection over Union (IoU) with the ground-truth information greater than 30\%; then, for all matches, the classifications induced by the classifier scores (the class corresponding the the maximum classification score) were compared with the ground-truth, with and without the calibration performed by the Foveal Observation model; the entropy of the classification scores was also compared, in both calibration cases.
    
    \item Data Fusion techniques: for each map cell $(x,y)$, there might be none to several detections at a given time step $t$; Let ${\cal I}_t^{x,y}$ be the set of detections at instant $t$ whose bounding boxes intersect the map cell $(x,y)$; at each time step $t$, the map cell $(x,y)$ update is repeated for every detection belonging to the set ${\cal I}_t^{x,y}$; the calibrated classification scores by the Foveal Observation Model (\ref{eq:obsModel_scores}) serve as input to the four fusion techniques: Na\"{i}ve Bayes update (\ref{eq:prod_rule}), Kaplan Update, Modified Kaplan's rule (\ref{eq:mod_kaplan}) and sum rule (\ref{eq:naive_update}); the next gaze direction was determined randomly instead of through Active Perception methods, in order to evaluate the raw performance of the fusion methods without considering gaze selection; the mean of the accuracy metrics across images was computed.
    
    \item Active Perception techniques: the performance of the different acquisition functions was compared; each of the 50 images has been foveated 10 times, with the foveation points having been chosen by the various active perception techniques; the initial focal point was randomly chosen for each image, being the same for every technique; the detections at each iteration were used to update the information of the map using the Modified Kaplan update (\ref{eq:mod_kaplan}).
    
\end{enumerate}

\subsection{Experimental Results and Discussion}

\subsubsection{Validation of the Foveal Observation Model}
\begin{figure*}[h]
\centering
\subfigure[Accuracy comparison.]{\includegraphics[width=0.35\linewidth]{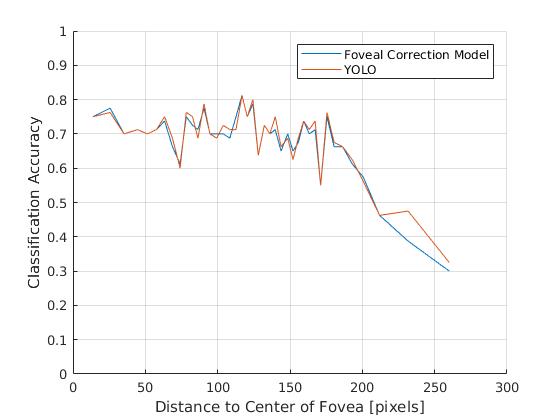}\label{fig:accuracyVSdistance}}
\subfigure[Entropy comparison.]{\includegraphics[width=0.35\linewidth]{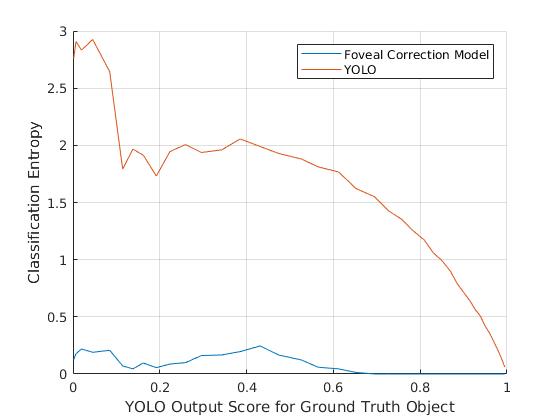}\label{fig:entropyVsScore}}
\caption{Classification entropy employing (blue) and without employing (red) the Foveal Observation Model.}
\label{fig:classif_entr_fom}
\end{figure*}

In Fig.~\ref{fig:accuracyVSdistance}, the comparison of the classification performance with and without the Foveal Observation Model is illustrated, with the metrics values resulting from averaging over all the images. The accuracy lines in Fig. \ref{fig:accuracyVSdistance} are very similar, meaning that modelling the detections with the observation model does not generally lead to a drop in detection performance, proving the validity of the Foveal Observation Model. On the other hand, detection uncertainty is vastly improved, as shown in Fig. \ref{fig:entropyVsScore}: low confidence scores output by the detector (i.e., detections significantly affected by the blur on the peripheries) typically have a high degree of entropy; when these scores are calibrated by the Foveal Observation Model, the entropy of the class confidence scores is much lower.

This is because the Foveal Observation Model tries to find the object class that better fits the distribution of the scores, accounting for the distance to the focal point (blur imposed by the foveation process), even if there is a big confusion among some of the classes, to present a more certain classification. Although this does not constitute an improvement in a one-step classification approach, as the lines in Fig. \ref{fig:accuracyVSdistance} are quite similar, it will still be useful for the multi-step classification approach that we are taking.

\subsubsection{Validation of the Fusion techniques}

\begin{figure}[htbp]
\centering
\includegraphics[width=0.66\linewidth]{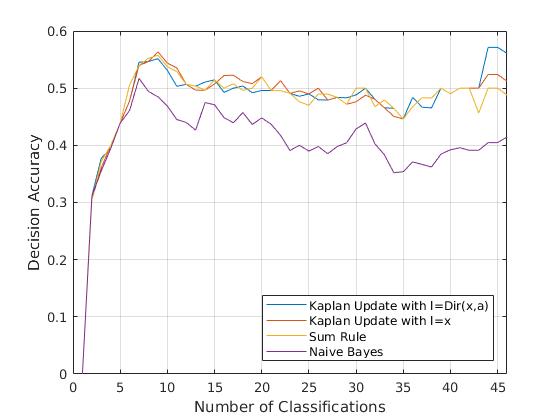}
\caption{Accuracy versus number classifications at each map location, for the fusion techniques considered.}
\label{fig:accuracy_avg}
\end{figure}

In Fig. \ref{fig:accuracy_avg}, one can see how the average accuracy evolves as new bounding boxes are detected (each bounding box corresponds to one classification). It is possible to note that every algorithm achieves a similar performance on the accuracy, except for the Na\"{i}ve Bayes, where the performance is lower. Also, due to the drastic entropy reduction imposed by the Foveal Observation Model, the Na\"{i}ve Bayes approach was greatly affected by most of the classes having a score closer to zero.

\subsubsection{Validation of the Active Perception techniques}
\begin{figure}[h]
\centering
\includegraphics[width=0.66\linewidth]{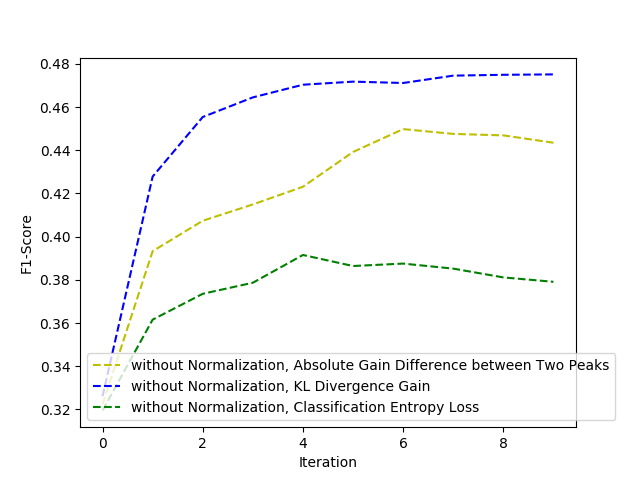}
\caption{Comparison between the F1-Scores using the three different acquisition functions.}
\label{fig:acquisition_perform}
\end{figure}

\begin{figure}[h]
\centering
\includegraphics[width=0.66\linewidth]{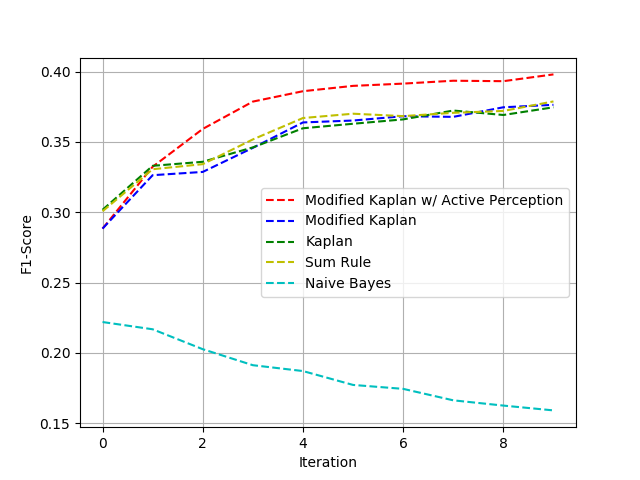}
\caption{F1-Score comparison of the Modified Kaplan using the acquisition function "KL Divergence Gain" (red), against the ones choosing the focal point randomly.}
\label{fig:f1_score_active}
\end{figure}

In Fig.~\ref{fig:acquisition_perform}, one can observe that the KL Divergence is the acquisition function achieving the highest F1-score. In Fig.~\ref{fig:f1_score_active}, active gaze selection is compared with random gaze selection. The experiment is the same as before, but now the Modified Kaplan fusion method with KL divergence acquisition function is tested against all the other fusion methods which choose the next focal point randomly. One can immediately notice that the Active Perception method achieves higher F1-Score at almost every iteration than the other fusion algorithms with random gaze selection.

Another important aspect is the growth rate of the performance on classifying the objects on the image. Since the goal is to find and classify every object on the image in the least number of gaze shifts, analysing how fast the algorithm can detect and correctly classify most of the objects is a key factor. As one can see, choosing the next focal point by maximizing the predicted gain on the average KL divergence of the map, achieves an F1-Score around the third iteration that can not be surpassed by any of the methods that use random search.

Besides the improved growth rate, we can also see in Fig. \ref{fig:f1_score_active} that choosing the next focal point by maximizing the KL Divergence Gain contributes to an overall performance improvement (on average) of around 2-3\% on the F1-Scores after the 10 iterations of the experiment.

\section{Conclusions}
\label{sec:concl}

In this work, we propose a methodology integrating a simulated foveal sensor, an object detector, an observation model, and data fusion and active perception techniques to perform visual exploration of a scene, in order to locate and correctly classify the objects in an image with the least number of gaze shifts. The main contribution of the paper is the Foveal Observation Model, which calibrates the scores of the Deep Object Detector to the foveal image topology in a fast training process without the need to retrain the deep object detector itself. We have shown that this calibration process significantly reduces the entropy of the classification score vectors. We embed the Foveal Observation model in several data fusion processes an evaluate performance both in random search and with active search. The proposed method is able to improve the F1-scores by 2-3 pp after a few gaze shifts, while the active perception method also achieves a score higher than the best achieved when randomly searching.

\section*{Acknowledgments}

This work was supported by FCT through projects LARSyS-UIDB/50009/2020, HAVATAR-PTDC/EEI-ROB/1155/2020, and SHIFTHRI-CMU/TIC/0026/2021.

\bibliographystyle{ieeetr}
\bibliography{icdl_paper_64}

\end{document}